\documentclass[10pt,twocolumn,letterpaper]{article} 

\usepackage{avss}
\usepackage{times}
\usepackage{epsfig}
\usepackage{graphicx}
\usepackage{amsmath}
\usepackage{amssymb}

\usepackage{hyperref}
\usepackage{enumitem}
\usepackage{eso-pic}

\setlist[itemize]{nosep, topsep=0pt, partopsep=0pt, leftmargin=*}

\avssfinalcopy 

\def\avssPaperID{36} 

\ifavssfinal\fi
\begin{document}

\title{ATL-Diff: Audio-Driven Talking Head Generation with Early Landmarks-Guide Noise Diffusion}

\author{
    Son Thanh-Hoang Vo\textsuperscript{1}, 
    Quang-Vinh Nguyen\textsuperscript{1}, 
    Seungwon Kim \textsuperscript{1},
    Hyung-Jeong Yang \textsuperscript{1}, \\
    Soonja Yeom\textsuperscript{2}, 
    and Soo-Hyung Kim\textsuperscript{1}\textdagger\\
    \textsuperscript{1}\textit{Chonnam National University}, Gwangju, Republic of Korea \\
    \textsuperscript{2}\textit{University of Tasmania}, Hobart, Australia \\
    {\tt\small \{hoangsonvothanh, vinhbn28, seungwon.kim, hjyang\}@jnu.ac.kr} \\
    {\tt\small soonja.yeom@utas.edu.au} \\
     \textsuperscript{\textdagger}Corresponding author: {\tt\small shkim@jnu.ac.kr}
}

\maketitle


\begin{abstract}
Audio-driven talking head generation requires precise synchronization between facial animations and audio signals. This paper introduces ATL-Diff, a novel approach addressing synchronization limitations while reducing noise and computational costs. Our framework features three key components: a Landmark Generation Module converting audio to facial landmarks, a Landmarks-Guide Noise approach that decouples audio by distributing noise according to landmarks, and a 3D Identity Diffusion network preserving identity characteristics. Experiments on MEAD and CREMA-D datasets demonstrate that ATL-Diff outperforms state-of-the-art methods across all metrics. Our approach achieves near real-time processing with high-quality animations, computational efficiency, and exceptional preservation of facial nuances. This advancement offers promising applications for virtual assistants, education, medical communication, and digital platforms. The source code is available at: \href{https://github.com/sonvth/ATL-Diff}{https://github.com/sonvth/ATL-Diff}
\end{abstract}


\section{Introduction}
Audio-driven talking head generation aims to synthesize realistic facial animations synchronized with speech from a still identity image. This technology has gained increasing interest due to its broad applicability, including virtual assistants, education, healthcare, entertainment, and immersive platforms like the Metaverse \cite{pataranutaporn2021ai, christoff2023application}.

Conventional approaches typically extract motion-related features from audio and embed them directly into generative models \cite{du2023dae}. Some methods employ facial landmarks or segmentation as intermediate representations before image synthesis \cite{wang2020mead, 10.1145/3664647.3681108}. Although these approaches have shown promising results, they still face challenges in achieving precise lip-audio synchronization and preserving facial identity. Direct audio-to-image generation often introduces noise, which hinders visual quality and increases computational cost.

Addressing the challenges in Audio-driven Talking Face Generation is the primary objective of this paper. This paper introduces \textbf{A}udio-Driven \textbf{T}alking Head Generation with Early \textbf{L}andmark-Guided Noise \textbf{Diff}usion (ATL-Diff), an innovative approach that isolates the audio signal to prevent interference during the general process. The main contributions of this paper can be summarized as follows:
\begin{itemize}
\item Proposed a new method for audio-driven talking head generation problem, achieving superior results compared to other models.
\item Build a sequence of milestones that corresponds to the video frame correctly from audio signal using the \textit{Landmarks Generation Module} module (section \ref{sec:propose_1}) as an intermediary step. This is demonstrated in Table \ref{tab:ablationstudies1}.
\item We leverage the power of the Diffusion model while introducing an entirely innovate noise distribution approach called \textit{Landmark} (section\ref{sec:propose_2}). In this method, the landmark sequence directs the noise distribution within the Diffusion model. This approach significantly reduces inference costs while maintaining high output quality. Table \ref{tab:cost} demonstrates how this contributes to improvements compared to other methods. 
\item Additionally, we refine the Diffusion model to support identity reconstruction, ensuring the preservation of facial characteristics (section \ref{sec:propose_3}). 
\end{itemize}

\begin{figure*}[t]
\centerline{\includegraphics[width=0.90\linewidth]{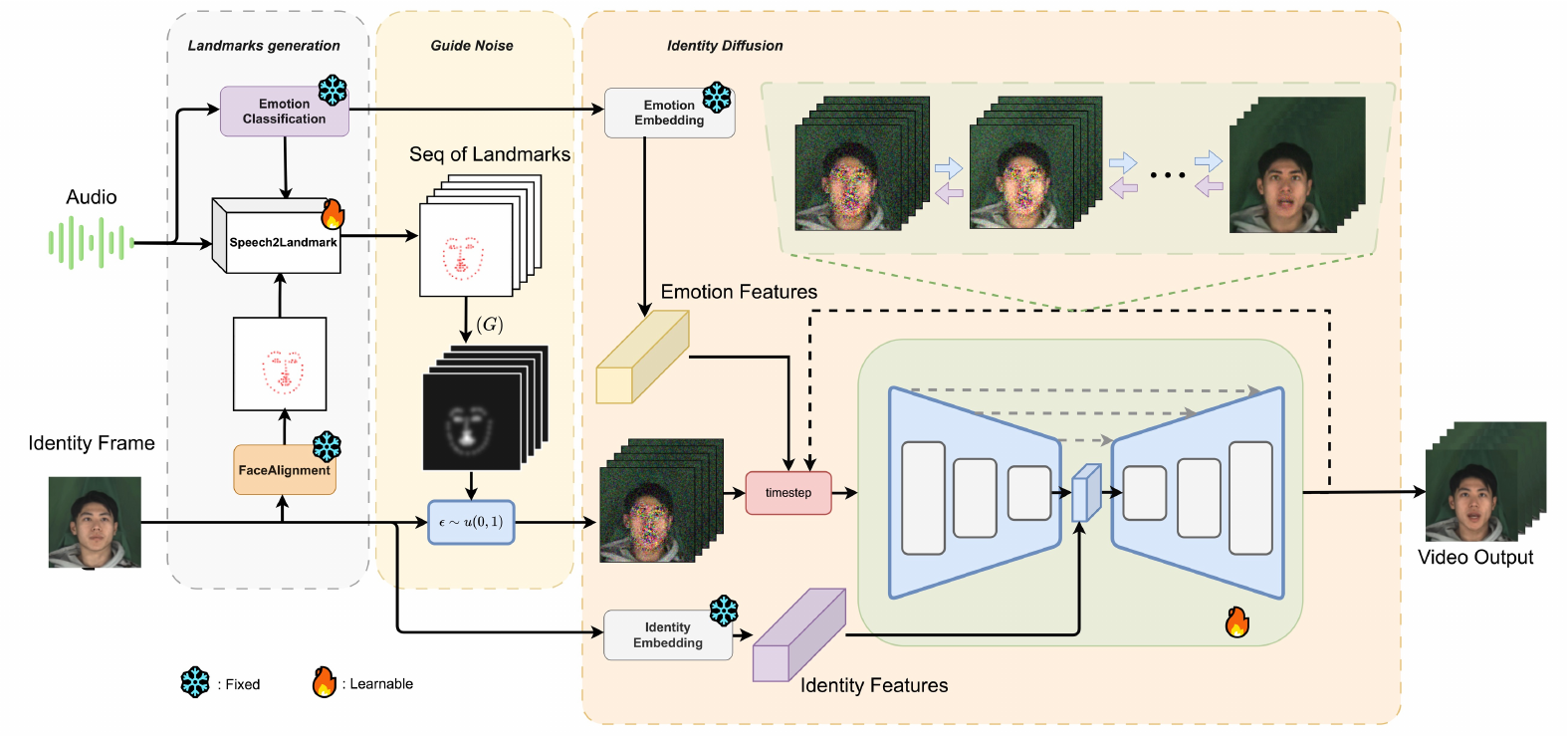}}
\caption{ATL-Diff architecture with three key components: Landmarks Generation Module, Landmarks-Guide Noise approach, and 3D Identity Diffusion. Snowflake symbols represent pre-trained models, while fire symbols indicate trainable models.}
\label{fig:Overview}
\end{figure*}

\section{Proposed Method}
\textbf{Overview}: The audio-driven talking head generation problem takes two inputs: an audio signal and an identity image, and will generate a video of the subject speaking. Figure \ref{fig:Overview} shows our proposed model with three main components transforming audio and identity inputs into synchronized facial animations.

\subsection{Landmarks Generation Module}
\label{sec:propose_1}
\begin{figure}[b]
\centerline{\includegraphics[width=\linewidth]{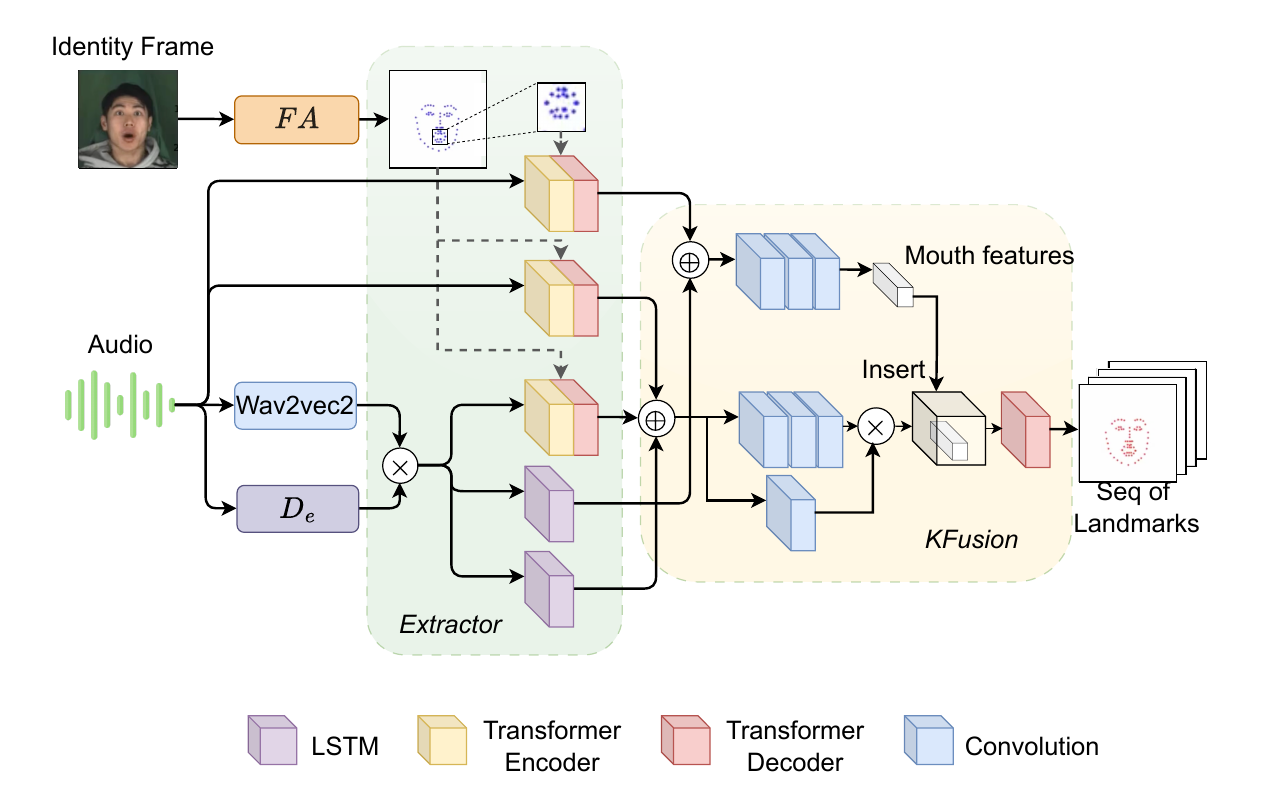}}
\caption{Speech to Landmark module architecture in the Landmarks Generation process. The Global Domain (first two paths) extracts features from raw audio, while the remaining paths form the Context Domain.}
\label{fig:LandmarksGeneration}
\end{figure}

This module was developed with the goal of constructing a sequence of landmarks from audio signals, drawing inspiration from dual-domain learning and features fusion techniques to achieve highly comprehensive results \cite{10222602}. Regarding Figure \ref{fig:LandmarksGeneration}, this module is divided into two components, The first component, called the Extractor, is designed to learn the correlation between audio signals and landmarks. The second component, KFusion integrates information from both domains and reconstructs it into a sequence of landmarks.

In the \textbf{Extractor} component, we define two complementary domains: the Global Domain and the Context Domain. In the Global Domain, two Transformer Encoder blocks \cite{vaswani2017attention}, denoted as \( \mathbf{E}_f \) and \( \mathbf{E}_m \), extract features from the raw audio signal \(x\) along two distinct branches. These features are then fused with facial landmarks \(v\) and \(v_m\) (full face and mouth region, respectively), extracted from the identity image using a facial alignment model \cite{bulat2017far}, and processed via a Transformer Decoder \( (\mathbf{D}) \). This design enables the model to focus more effectively on speech-related facial motions. The process is formulated as:
\begin{equation}
\begin{split}
g_{f} &= \mathbf{E}_f(\texttt{Conv}(\mathbf{D}(x), v)) \\
g_{m} &= \mathbf{E}_m(\texttt{Conv}(\mathbf{D}(x), v_m))
\end{split}
\label{eq:GloablDomain}
\end{equation}

In parallel, the Context Domain leverages pretrained models to extract higher-level and emotion-aware features. Specifically, we use wav2vec2.0 \cite{baevski2020wav2vec} to extract robust audio representations and Conformer \cite{gulati20_interspeech} as an emotion recognition backbone \( (\mathbf{E}_e)\). These features are fused via element-wise multiplication to obtain the contextual representation \( c \), as shown in Equation~\ref{eq:ContentDomain}:

\begin{equation}
c = \mathbf{E}_e(x) \times \texttt{wav2vec}(x)
\label{eq:ContentDomain}
\end{equation}

This fusion enables the model to incorporate both phonetic and emotional cues, thereby enhancing the quality and expressiveness of the predicted facial motion.

One branch has the structure of the Global Domain to learn the correlation between facial features and emotional features \((c_{ef})\). Two LSTM modules are used to learn and refactor features represented as \((c_m)\) and \((c_f)\) for the next phase.

\textbf{KFusion}: This component integrates features directly from the outputs of the Extractor and reconstructs them to return a sequence of landmarks. Since the features are divided into two parts whole facial and mouth area. This component uses concatenation to combine these features into two primary sets:  \(x_f\) for the whole face and \(x_m\) for the mouth area, Equation \ref{eq:KFusion1} represents this process. 
\begin{equation}
\begin{split}
    x_m = c_m \oplus g_m \quad ; \quad x_f = c_f \oplus c_{ef} \oplus g_f 
\end{split}
\label{eq:KFusion1}
\end{equation}
Each obtained feature is processed through a block consisting of \(Conv(k=1) \rightarrow Conv(k=3) \rightarrow Conv(k=1) \) denoted as \(Conv\) for simplicity. Inspired by the Bottleneck Block from \cite{liu2024kan}, a Residual path is incorporated for the facial feature map, creating a block denoted as \(RConv\). Next, the mouth features are inserted into the facial features at the corresponding position \((L_m)\), as shown in Figure \ref{fig:LandmarksGeneration}. Finally, a KAN network is used to predict the landmarks \(\hat{y}_{lm}\). The formula for this process is presented in Equation \ref{eq:KFusion2}.
\begin{equation}
\begin{split}
    x'_f &= RConv(x_f) \times Conv(x_f) \\
    x'_f &[:,:,L_m] \leftarrow Conv(x_m) \\
    \hat{y}_{lm} &= KAN(x'_f)
\end{split}
\label{eq:KFusion2}
\end{equation}
\subsection{Landmarks-Guide Noise Approach}

\label{sec:propose_2}
The proposed method employs a Diffusion model for the video generation. As phase and as mentioned, an approach is required to isolate the audio signal from the generation model. To achieve this, this component introduces movement information through the sequence of landmarks extracted in the previous step. This approach is inspired by the Inpainting method \cite{lugmayr2022repaint}. The input to the Diffusion model will have noise added according to a distribution guided by a mask layer, which is built from the sequence of landmarks.
First, we transform the landmarks matrix into an image of size \(128 \times 128\), setting the values at the landmark positions to 1. Gaussian blur is then applied to the newly created image, resulting in a mask with higher values around the landmark points, which gradually decrease as the distance from these points increases. Next, random noise is generated using a uniform distribution within the range \(\eta \in [0,1]\). Since the guide mask is distributed using Gaussian, the noise must also be uniformly distributed. When adding noise to the Identity Frame according to the mask, noise regions will appear at the landmark points, with a radius \(r = \frac{k-1}{2}\), where k is an odd number representing the kernel size. This entire process is described by Equation \ref{eq:LandmarkGuideNoise}.
Here \(\sigma\) represents the standard deviation, and \(\delta\) denotes the minimum noise threshold for the input. Some noise must be applied to the entire image to ensure the Diffusion model can generate detailed features without oversampling.
\begin{equation}
\begin{split}
   G(x,y) &= \frac{1}{2\pi \sigma^2} \epsilon^{-\frac{x^2 + y^2}{2\sigma^2}} \\
   I'(x,y) = \sum^{k}_{i=-k} &\sum^{k}_{j=-k} I(x+i, y+j) \cdot G(i,j) \\
   \eta \sim   U[0,1] \quad &; \quad \hat{I} = (\delta + I' * \eta)
\end{split}
\label{eq:LandmarkGuideNoise}
\end{equation}

\subsection{3D Identity Diffusion}
\label{sec:propose_3}
After generating the sequence of noise input frames, we utilized a 3D Diffusion network for the denoising task, producing a sequence of high-quality images. As the noise regions are concentrated around areas requiring motion, the model simultaneously reconstructs these movements during the denoising process. Additionally, the use of the 3D Diffusion model ensures that transitions between frames are reconstructed as smoothly as possible.

The model incorporates two optimization features: an emotion feature (\(w_e\)) extracted from audio as described in Section \ref{sec:propose_1}, and identity features (\(w_i\)) from the Identity Image via a pre-trained ResNet. The emotion feature \(w_e\) combines with the timestep vector through matrix addition, ensuring consistent emotion recreation even with reduced timesteps. Meanwhile, \(w_i\) is incorporated after the Downsampling stage, where multiplying latent features with \(w_i\) amplifies identity-specific information for accurate reconstruction during Upsampling.

The 3D U-Net used in this paper consists of three 3D Residual Blocks with channel sizes of [32, 64, 128]. The dimension of \( w_i \) is \( B \times 2048 \), which is reshaped to \( B \times 32 \times 64 \). The output size matches the input, which is \( B \times F \times 128 \times 128 \times 3 \), where \( B \) is the Batch size and \( F \) is the number of frames. For the experiments in this paper, \( F \) is set to 30. The output is a sequence of frames depicting the Identity moving across frames.

\section{Experiments and Results}

\begin{table*}[h]
\centering
\caption{Performance comparison with state-of-the-art methods. Bold: best results. ↑/↓: higher/lower is better}
\label{tab:quantitative}

\begin{tabular}{lccccc|ccccc}
\textbf{Dataset} & \multicolumn{5}{c}{\textbf{CREMA-D}} & \multicolumn{5}{c}{\textbf{MEAD}} \\
\hline
\textbf{Method} & PSNR \(\uparrow\) & SSIM \(\uparrow\) & FID \(\downarrow\) & M-LMD \(\downarrow\) & LMD \(\downarrow\) & PSNR \(\uparrow\) & SSIM \(\uparrow\) & FID \(\downarrow\) & M-LMD \(\downarrow\) & LMD \(\downarrow\) \\
\hline
MEAD \cite{wang2020mead} & 20.915 & 0.703 & 78.835 & 8.569 & 11.092 & 28.554 & 0.683 & 22.521 & 2.525 & 3.162 \\
MakeItTalk \cite{zhou2020makelttalk} & 23.357 & 0.758 & 33.543 & 3.482 & 3.982 & 22.350 & 0.819 & 58.886 & 3.748 & 3.980 \\
EVP \cite{ji2021audio} & 27.988 & 0.687 & 8.872 & 2.764 & 3.284 & 28.133 & 0.773 & 13.434 & 2.055 & 2.178 \\
Diff-Heads \cite{stypulkowski2024diffused} & 28.545 & 0.819 & 12.433 & 1.327 & 1.487 & 27.373 & 0.810 & 18.320 & 3.087 & 3.281 \\
\textbf{Ours} & \textbf{29.055} & \textbf{0.985} & \textbf{7.995} & \textbf{0.701} & \textbf{0.840} & \textbf{28.883} & \textbf{0.944} & \textbf{8.842} & \textbf{1.574} & \textbf{1.840} \\
\hline
\end{tabular}
\end{table*}

\begin{figure*}[h]
\centerline{\includegraphics[width=0.80\linewidth]{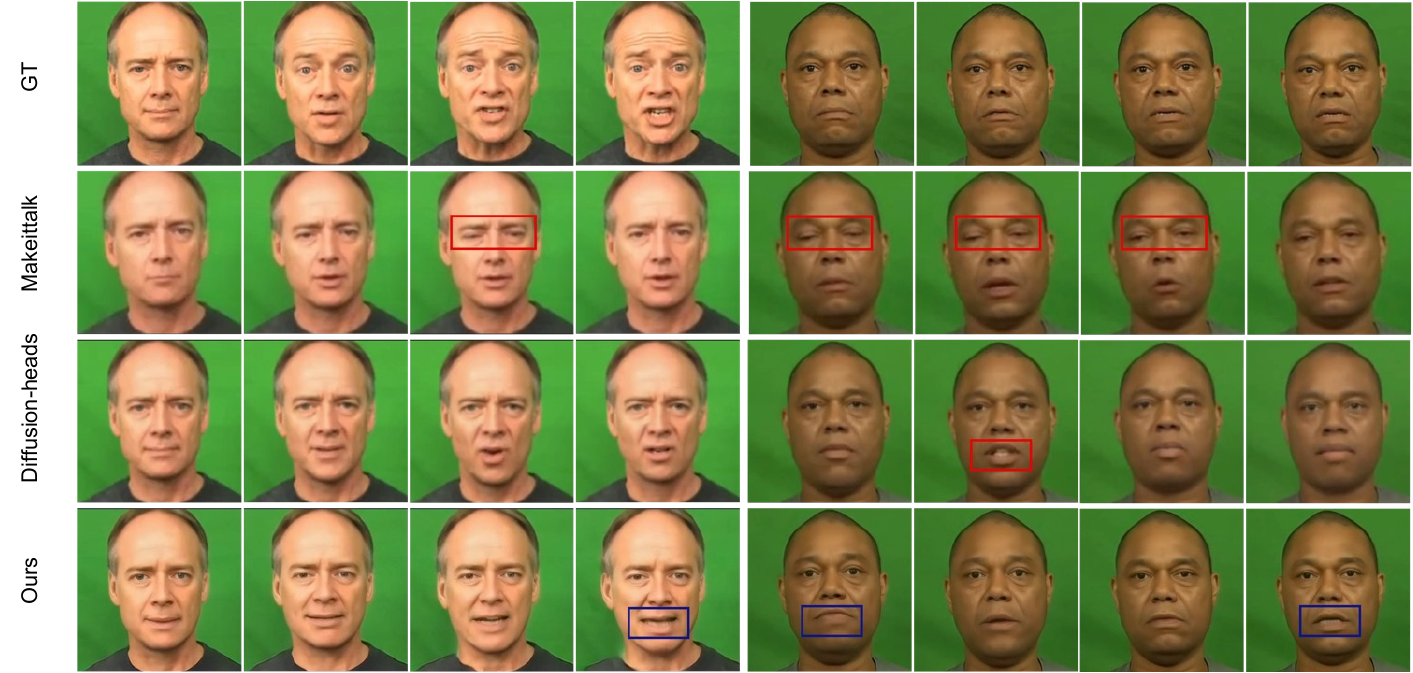}}
\caption{Qualitative comparisons using CREMA samples 1101 and 1083. Red box: previous methods' eye detail limitations; blue box: our improved mouth movement reconstruction.}
\label{fig:Quantitative results}
\end{figure*}

\subsection{Dataset}
We used MEAD \cite{wang2020mead} and CREMA-D \cite{6849440} datasets for training and testing. MEAD contains ~40 hours of audiovisual clips per person across 60 individuals with 8 emotions at 3 intensity levels. CREMA-D includes 7442 samples from 91 actors with 6 emotions. Videos were resized to 128×128 pixels, processed at 30 FPS, and split into 1-second clips. Audio was converted to 16 kHz mono format. Data was divided 90\% for training (of which 10\% for validation) and 10\% for testing.

\subsection{Experimental setup}
Experiments ran for 300 epochs using MSE loss and Adam optimizer (learning rate 1e-4 with Cosine Annealing \cite{loshchilov2017sgdr} to 1e-6). For the Diffusion model, DDIM \cite{song2020denoising} scheduler with 1000 training steps and 8 inference steps was used. All experiments were conducted on an NVIDIA RTX 4090.

\subsection{Results}
The models used for comparison were open-source, with publicly available code and pre-trained weights. We implemented and utilized these models to generate the results to maintain consistency all inference processes were performed on the same device.

\subsubsection{Quantitative Results}
To assess the performance of the proposed model, we used five distinct evaluation metrics. The Peak Signal-to-Noise Ratio (PSNR) and Structural Similarity Index Measure (SSIM) were utilized to assess the structural quality of the generated images in comparison to the ground truth. The Fréchet Inception Distance (FID) \cite{10.5555/3295222.3295408} metric was used to measure the similarity between the generated images and real images. Additionally, we applied Landmark Distance in the whole face (LMD) \cite{chen2018lip} and the mouth region (M-LMD) to evaluate the temporal consistency and correlation of the generated video motion with the ground truth sequences. These comprehensive metrics provided a thorough evaluation of both the visual fidelity and the motion accuracy of the proposed approach.

\begin{table}[b]
\centering
\caption{Inference time comparison per sample.}
\label{tab:cost}
\begin{tabular}{lcc}
\textbf{Method} & \textbf{FPS} & \textbf{Time (s)} \\ \hline
MakeItTalk \cite{zhou2020makelttalk} & 24 & 18.98 \\
EVP \cite{ji2021audio} & 25 & 20.45 \\
Diff-Heads \cite{stypulkowski2024diffused} (step = 50) & 25 & 68.32 \\
Diff-Heads \cite{stypulkowski2024diffused} (step = 8) & 25 & 13.08 \\
\textbf{Ours (step = 8)} & 30 & \textbf{2.02} \\ \hline
\end{tabular}
\end{table}
\textbf{Learning Ability:} As shown in Table \ref{tab:quantitative}, our model achieves the best performance across all evaluation metrics. Notably, the results for PSNR and SSIM demonstrate the effectiveness of leveraging the generative capabilities of Diffusion models. The high-quality image generation achieved by our model is further validated by its top-performing results, particularly in the generation phase using the Diffusion model \cite{stypulkowski2024diffused}. Furthermore, the LMD and M-LMD metrics highlight the effectiveness of our Landmark Generation module which successfully generates high-quality landmark sequences. These sequences enable the Diffusion model to accurately reconstruct the corresponding images.

In addition to visual and motion accuracy, we also highlight the efficiency of our proposed model in terms of inference speed. In this experiment using the CREMA-D dataset, where the average sample duration is 2 seconds, our model performs inference on videos at up to 30 frames per second (FPS). As shown in Table \ref{tab:cost}, the results demonstrate that our approach is highly efficient, with a processing delay of approximately 2 seconds. This indicates that the model can process a 1-second video segment in just 1 second, making it nearly suitable for real-time applications.

\subsubsection{Qualitative Results}
For qualitative analysis, we present visual comparisons between our method and others in Figure \ref{fig:Quantitative results}. Our model consistently demonstrates superior image quality, especially when compared to other approaches, such as \cite{zhou2020makelttalk}, whose outputs appear noticeably blurry and lack sharpness. Additionally, certain facial regions, such as the eyes and mouth, exhibit unnatural movements in the outputs of other methods, particularly when compared to the ground truth. 

For instance, in \cite{zhou2020makelttalk}, the eyes often appear closed, while in \cite{stypulkowski2024diffused}, the mouth movements are inaccurately timed. In contrast, our model excels at generating mouth movements that closely match the ground truth, accurately capturing both the opening and closing actions with remarkable precision. This is attributed to dedicated branch in our Landmark Generation module, which is specifically designed to learn mouth dynamics, effectively captured and rendered by the 3D Identity Diffusion model.

However, our approach has some limitations. It struggles to reproduce finer details, such as wrinkles on the forehead, and the lip shapes during articulation are not yet perfectly accurate.

\subsubsection{Ablation studies}
\begin{table}[h]
\centering
\caption{Impact of Landmark Generation module components on performance.}
\label{tab:ablationstudies1}
\begin{tabular}{lrr}
\textbf{Method} & \textbf{M-LMD \(\downarrow\)} & \textbf{LMD \(\downarrow\)} \\ \hline
w/o KFusion & 3.300 & 8.624 \\
w/o Content Domain & 2.799 & 5.490 \\
w/o Global Domain & 2.120 & 4.126 \\
with MLP & 1.380 & 3.990 \\
\textbf{with KAN} & \textbf{1.089} & \textbf{1.769} \\ \hline
\end{tabular}
\end{table}
We evaluated the impact of each domain and the performance of the KAN network within the Landmark Generation module. Specifically, we conducted experiments by disabling one of the domains and replacing the KAN network with an MLP. The results, presented in Table \ref{tab:ablationstudies1}, demonstrate that the model achieves the best performance when both domains are used simultaneously. Furthermore, the KAN network outperforms the MLP by a significant margin, highlighting its effectiveness for this task.

\subsubsection{Generalization}

\begin{figure}[b]
\centerline{\includegraphics[width=0.9\linewidth]{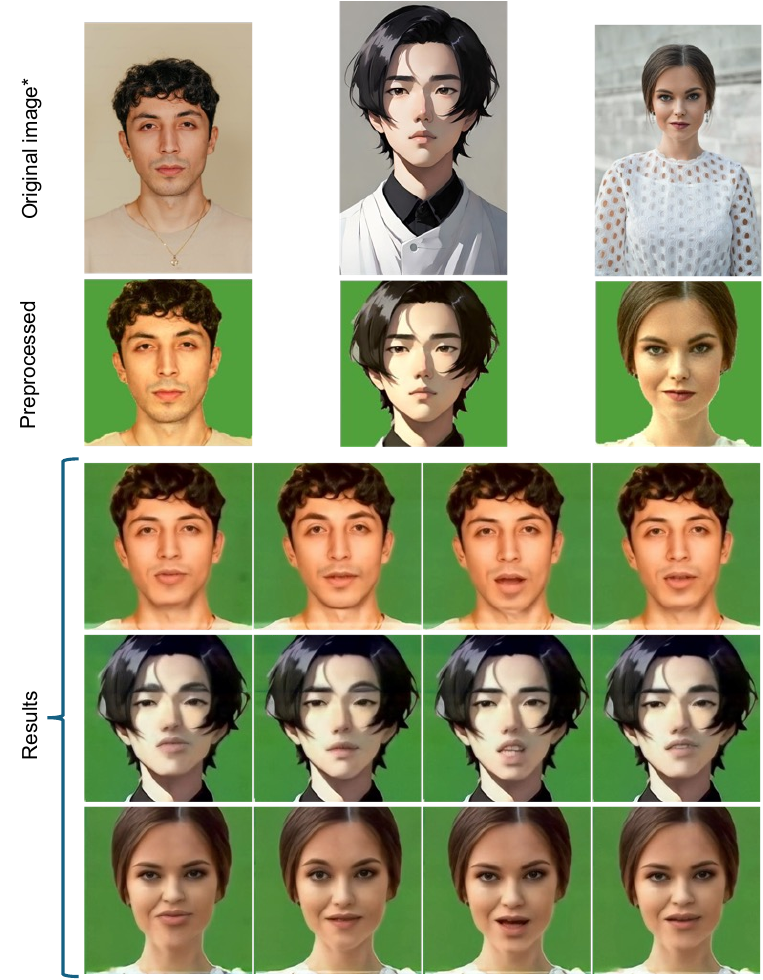}}
\caption{Results on real-world data. Top: original images; middle: preprocessed images; bottom: our generated results. (*Free license images)}
\label{fig:Opendata results}
\end{figure}

To evaluate the robustness of our proposed method, we tested its performance on real-world datasets sourced from the internet. This experiment was designed to demonstrate the capability of the model to generalize across diverse and potentially noisy inputs.

We collected free-licensed images from various online sources to represent a diverse range of facial features including men, women, and cartoon characters from different ages, and ethnicities. For the audio, we recorded our voices and using scripts provided by the dataset. The preprocessing steps involved face detection and segmentation using Mediapipe, followed by standardizing the images with a green screen background. Figure \ref{fig:Opendata results} illustrates these preprocessing steps, with the first and second rows showing the transformations. The audio recordings were resampled to 16000 Hz and converted to a mono-channel format to meet with the model's requirements, ensuring consistency across all inputs.

As shown in Figure \ref{fig:Opendata results}, our model performs exceptionally well in recreating Identity details and movements. Although pose variations have not yet been fully explored, the movements of the eyes and mouth regions are highly diverse, effectively capturing and representing unique identity characteristics.

\section{Conclusion}
This paper presents ATL-Diff, a novel approach for audio-driven talking head generation addressing key challenges in facial animation synthesis. Our three-component system: Landmarks Generation module, Landmarks-Guide Noise approach, and 3D Identity Diffusion network effectively preserves facial identity while ensuring precise audio synchronization. Experiments on CREMA-D and MEAD datasets demonstrate superior performance across all metrics compared to existing methods. The approach achieves near real-time processing speeds while maintaining high-quality output, making it suitable for practical applications.

\section*{Acknowledgement}
This work was supported by the Institute of Information \& Communications Technology Planning \& Evaluation (IITP) under the Artificial Intelligence Convergence Innovation Human Resources Development (IITP-2023-RS-2023-00256629) grant funded by the Korea government (MSIT), the ITRC (Information Technology Research Center) support program IITP-2025-RS-2024-00437718) supervised by IITP, the National Research Foundation of Korea(NRF) grant funded by the Korea government(MSIT) (RS-2023-00219107).



{\small
\bibliographystyle{ieee}
\bibliography{egbib}
}

\end{document}